\documentclass[11pt,a4paper]{article}
\usepackage{acl2014}
\usepackage{times}
\usepackage{latexsym}
\usepackage{url}
\usepackage{color}
\usepackage{multirow}
\usepackage{comment}
\usepackage{amsmath} 
\newtheorem{theorem}{Theorem}
\usepackage{amssymb}

\usepackage{etoolbox}
\usepackage{graphicx}
 
 \usepackage{enumitem}
\newenvironment{tightitemize}%
  {\begin{itemize}[topsep=0pt, partopsep=0pt] %
    \setlength{\itemsep}{0pt}%
    \setlength{\parskip}{0pt}%
    }%
  {\end{itemize}}
  
  \usepackage{color}
  {\begin{enumerate}≈ \footnotesize%
    \setlength{\itemsep}{0pt}%
    \setlength{\parskip}{0pt}%
    }%
  {\end{enumerate}}
  
  \usepackage{CJKutf8}

\usepackage{soul}
\title{The NLP Engine: A Universal Turing Machine for NLP}
\author{Jiwei Li$^1$ and Eduard Hovy$^2$
\\$^1$Computer Science Department, Stanford University, Stanford, CA 94305\\
$^2$Language Technology Institute, Carnegie Mellon University, Pittsburgh, PA 15213\\jiweil@stanford.edu ~~~~~~~ehovy@andrew.cmu.edu
}

\newtheorem{Assumption}{Assumption}[theorem]
\newtheorem{Requirement}{Requirement}[theorem]

\begin{document}
\maketitle
\begin{abstract}
It is commonly accepted that machine translation is a more complex task than part of speech tagging.  But how much more complex? 
In this paper we make an attempt to develop a general framework and methodology for computing the informational and/or processing 
complexity of NLP applications and tasks.  
We define a universal framework akin to a Turning Machine that attempts to fit (most) NLP tasks into one paradigm. 
We calculate the complexities of various NLP tasks using measures of Shannon Entropy,
and compare `simple' ones such as part of speech tagging to `complex' ones such as machine translation.  
This paper provides a first, though far from perfect, attempt to quantify NLP tasks under a uniform paradigm.
We point out current deficiencies and suggest some avenues for fruitful research.  
\end{abstract}

\section{Introduction} 

The purpose of this paper is to suggest a unified framework in which modern NLP research can quantitatively describe and compare NLP tasks. 
Even though everyone agrees that some NLP tasks are more complex than others, 
e.g., machine translation is `harder' than syntactic parsing, which in turn is `harder' than part-of-speech tagging, 
we cannot compute the relative complexities of different NLP tasks and subtasks. 

In the typical current NLP paradigm, researchers apply several machine learning algorithms to a problem, report on their performance levels, 
and establish the winner as setting the level to beat in the future.  
We have no single overall model of NLP that subsumes and regularizes its various tasks.  
If you were to ask NLP researchers today they would say that no such model is possible, and that NLP is a collection of several semi-independent 
research directions that all focus on language and mostly use machine learning techniques.  
Researchers will tell you that a good summarization system 
on DUC/TAC dataset
obtains a ROUGE score of 0.40, a good French-English translation system 
achieves a BLUE score of 37.0, 20-news classifiers can achieve accuracy of 0.85, and named entity recognition systems a recall of 0.95, 
and these numbers are not comparable. 
Further, we usually pay little attention to additional important factors such as  the performance curve with respect to the amount of training data, 
the amount of preprocessing required, the size and complexity of auxiliary information required, etc.  
And even when some studies do report such numbers, in NLP we don't know how to characterize these aspects in general and across applications, how to quantify them in relationship to each other. 

We here describe our first attempt to develop a single generic high-level model of NLP.  
We adopt the model of a universal machine, akin to a Turing Machine but specific to the concerns of language processing, 
and show how it can be instantiated in different ways for different applications.  
We employ Shannon Entropy within the machine to measure the complexity of each NLP task. 

In his epoch-making work, Shannon \shortcite{shannon1951prediction} demonstrated how to compute the amount of information in a message.  
He considered the case in which a string of input symbols is considered one by one, and the uncertainty of the next is measured by counting how difficult it is to guess.  
We make the fundamental assumption that most NLP tasks can be viewed as transformations of notation, in which a stream of input symbols is transformed and/or embellished into a stream of output symbols (for example, POS tagging is the task of embellishing each symbol with its tag, and MT is the task of outputting the appropriate translation word(s)).  
Under this assumption one can ask: how much uncertainty is there in making the embellishment or transformation?  
This clearly depends on the precise nature of the task, on the associated auxiliary knowledge resources, and on the actual algorithm employed.  
We discuss each of these issues below.  
We first describe the key challenge involved in performing uncertainty comparison using the Entropy measure in Section 2. 
In Section 3 we provide high-level comments on what properties a framework should have to enable fair complexity comparison.  
In Section 4, based on the properties identified in Section 3, we consider the theoretical nature of NLP tasks and provide suggestions for instantiating the paradigm. 
The framework is described in Sections 5, 6 and our results are presented in Section 7. 
We point out current deficiencies and suggest avenues for fruitful research in Section 8, followed by a conclusion.

\section{The Dilemma for Shannon Entropy}

\subsection{Review of Entropy and Cross Entropy}

Entropy, denoted as $-\sum_x p(y)\log(y)$, illustrates the  amount of information contained in a message, and can be characterized as the uncertainty of a random variable of a process. 
For example,  Shannon \shortcite{shannon1951prediction} reported an upper bound of 1.3 bits/character symbol for English character prediction and 5.9 bits/word symbol for English word prediction, meaning that it is highly likely that English word prediction is a harder task than English character prediction. 

If the output $Y^n=\{ y_0, y_{1}, ...y_n\}$ is a sequence generated from the input, a stationary stochastic process.  
Then the entropy of Y is given by:
\begin{equation}
H(Y)=\lim_{n\rightarrow\infty}H(y_n|y_{n-1}, y _{n-2} ..., )
\end{equation}
By the Shannon-McMillan-Breiman theorem \cite{algoet1988sandwich} this can be written as:
\begin{equation}
H(Y)=\lim_{n\rightarrow\infty}-\frac{1}{n}\log P(y_1,y_2,...,y_n)
\end{equation}
So we can define its hardness or complexity by computing entropy from the distribution $P(Y)$ for tasks like Shannon's word prediction model, 
or extend it to a noisy channel model \cite{shannon2001mathematical}: 
given a sequence of inputs $X$, the uncertainty of the output transformation is given by $H(Y|X)$, interpreted as the amount of uncertainty remaining 
about Y when X is already known. 

The true distribution
over
 $Y$ is hard to estimate.
Normally we estimate the upper bound of entropy --- the cross entropy denoted as $H(P, \hat{P})$----to approximate the true value of entropy:
\begin{equation}
H(P, \hat{P})=H(P)+D_{KL}(P||\hat{P})\geq H(P)
\end{equation}
where $D_{KL}(P||\hat{P})$ denotes the KL divergence between two distributions $P$ and $\hat{P}$. 
A good model will closely approximate $P$ using $\hat{P}$, leading to smaller value of $D_{KL}(P||\hat{P})$, i.e., bringing the value cross-entropy closer to that of the real one. 
Different models would obtain different values of $H(P, \hat{P})$.  \
Various studies since Shannon's work (e.g.,\cite{kucera1967computational,cover1978convergent,gopinath1987open,brown1992estimate})
have explored methods to lower the upper bound of character prediction entropy in English by using more sophisticated models. 

\subsection{The Dilemma for Entropy}

While entropy describes the intrinsic nature of the problem or task, its actual value estimation has to be determined by the specific model you adopt for prediction.
When Shannon approached the character prediction task, his wife acted as the predictor. 
Alternatively,
if Shannon had used a child as predictor, he would have obtained a much larger estimated entropy. 

Similarly, if one wishes to compare the entropy of two tasks,
for example,
to determine which language sequence is harder to predict, English or French, it would be problematic if one compares the entropy computed via a linguist for English and a child for French.  One requires twins who are mathematically and linguistically identical in terms of English and French for a fair comparison  \cite{cover2012elements}.  
However, in real world, it is almost impossible to find such twins. 
Different models are attuned differently to different scenarios, tasks, datasets, evaluation metrics, parameter settings, or optimization strategies. 
One model might not fit all tasks equally well, e.g., SVMs are not designed to predict probabilities, CRFs offer more insights in sequence labeling tasks than SVMs but are hard to use straightforwardly for text classification, etc. 

In summary, though entropy provides a theoretical definition about the uncertainty of a data source or task, the fact that its estimation must be performed using a real  specific model poses a dilemma for the accurate estimation of the uncertainty of tasks and hence for their fair comparison. 
\section{Prerequisites for Fair Comparison}

We claim that a framework should incorporate the following elements to enable a fair complexity comparison of disparate NLP tasks and systems: 

\paragraph{A universal measure:} Complexity can be measured in terms of multiple aspects (e.g., the amount of training data required, the amount of preprocessing required, the size and complexity of auxiliary information, training time, memory usage, or even lines of code).  
But we need a universal
and appropriate
 metric. 
In this work, we propose Shannon Entropy as the universal metric, which we believe reflects the intrinsic randomness, predictability, and uncertainty of datasets and tasks.  All the above aspects are highly correlated with Shannon Entropy.

\paragraph{A universal  Engine:}  A POS tagging system makes decisions by selecting tags with highest probability while a summarization system selects the top-ranked sentences.  A fair comparison of complexity, however, requires a single general and unified engine to define all (at least most of) NLP tasks within the same framework.  The abovementioned notation transformation paradigm, elaborated in the following section, accommodates most NLP tasks.  

\paragraph{A universal  model:} We cannot fairly compare the entropy obtained from a logistic regression model on POS tags to that produced from a large framework of interdependent alignment, phrase extraction, decoding algorithms for machine translation.  
A unified model should work with predictions for all (or at least most) current NLP tasks, and should make relatively accurate predictions (a random guess model, for example, is general but would not be helpful). 
We propose in Section 6 a candidate model.

\section{The Nature of NLP}

In order to propose a single multi-purpose unified Engine for NLP one has to adopt a very general perspective.  When constructing an NL system one typically assembles a variety of components.  Some of them are active modules that instantiate algorithms and perform transformations.  Others are passive resources (like lexicons, probability tables, or rulesets) that support the former.  Active modules are sometimes built to produce passive ones.  It is important to differentiate the role of modules in a framework in order to properly estimate the overall complexity.  In this section we first categorize the primary roles of NLP (sub)systems and then postulate that modern NLP algorithms (largely) fall into three distinct complexity types. 

\subsection{Three Classes of NLP System} 

NLP systems generally perform one of the following three functions/roles: (i) research into aspects of the nature of language(s), (ii) application tasks and subtasks, and (iii) support algorithms.  The majority of NLP development today falls into the second and third classes.  

Research into language includes such studies as determining the Zipfian nature and the entropy of language, discovering changes in patterns of use over time and across geographic regions, identifying text genres by for example creating word and constituent distribution profiles, and so on.  

Application tasks include machine translation, information retrieval, speech recognition, natural language interpretation (both syntactic parsing and semantic analysis), information extraction, question answering, dialogue processing, text summarization, text (sentence and multi-sentence) generation, sentiment analysis / opinion mining, text mining / harvesting, and others.  Subtasks include part of speech tagging, chunking, coreference resolution, text segmentation, query analysis, bitext alignment, reference generation, profiling/characterization of language producers, and many others, as well as numerous resource creation tasks including building monolingual and bilingual lexicons, distributional semantic word profiles and embeddings, word sense lists, ontologies / taxonomies, word-sentiment lists, and many others.  

Support algorithms include a variety of generic procedures that are reused in many applications.  In addition to the classic Finite State / Augmented Transition Network technology from the 1960s and later, modern NLP usually works with the statistical properties of large collections of words, and modern support algorithms such as HMMs and others generally assign and use count-derived scores to [sets of] words, such as tf.idf, PMI, and others, or distribute probability across sets of labels, words or documents, such as Expectation Maximization, Probabilistic Graphical Models, Topic Modeling algorithms, certain clustering algorithms, etc.  Some support algorithms focus on processing human labeling (annotation) and comparing the results of various different labeling agents (human and machine), such as Jensen-Shannon and other distribution comparison scoring, annotation optimization procedures, etc.  

\subsection{Three Levels of NLP Algorithm} 
\label{subsec:operations} 

We postulate that (almost every) NLP task / subtask can be defined as [a combination of] one of three basic operations, listed in order of complexity: 

{\bf Level 1: Prediction:} The algorithm reads its input, which includes principally a sequence of units of some kind, and predicts the next item in the sequence.  Example: predicting the next word in a stream, as used by Shannon to calculate the information content of text.  

{\bf Level 2: Labeling:} The algorithm reads its input and generate label(s) based on it.  Labeling tasks can be divided into two subcategories: 
\begin{itemize}\vspace*{-1mm}
\item {\bf Aligned Labeling}: there is a one-to-one correspondence between inputs and outputs.
Aligned tasks include most tagging tasks (e.g., named entity tagging and part-of-speech tagging). 
\item {\bf Unaligned Labeling}: no aligned correspondence exists between inputs and outputs.
Unaligned sequence-label tasks can be further divided into {\bf single-label} tasks such as categorization or clustering, in which a single label is assigned given the input (e.g., classification of documents each into one class), and {\bf sequence-label} tasks, in which a sequence of labels is produced (e.g., MT, where the labels are target language words).  
\end{itemize}

{\bf Level 3: Scoring:} The algorithm reads its input and assigns a score (without loss of generality, a real value between zero and unity) to some unit(s) in it.  The score may be a probability, rating, or some other score.  Example: tf.idf scoring of words.  

In a probabilistic paradigm, the probability of Level 1 tasks can be characterized as $P(Y)$ where $Y$ denotes the sequence to predict.  Level 2 tasks can be characterized as $P(Y|X)$ where $X$ denotes the input and $Y$ denotes the label(s) to generate.  

Often, one operation is used to perform another.  It is typical in modern-day (post-1990s) NLP to perform all kinds of labeling (Level 2 operations) by scoring all relevant possible categories (a Level 3 operation) and then returning the highest-scoring one as the selected tag.  This contrasts with pre-1990s NLP that generally computed a single result, such as the desired label, as the one and only possible answer.  

A task may require several operations in sequence.  For example, syntactic parsing requires labeling the part of speech tag of each word, labeling the start and end words of syntactic constituents, labeling the head of each constituent, and labeling the syntactic role of each constituent with regard to its immediate head.  Sometimes the label is drawn from a small set of possible tags that is predefined by theorists or the researcher, such as the part of speech tags.  Sometimes the label is provided in the text, such as the head word of a syntactic constituent.  Sometimes the label is a value computed by a scoring operation, such as the PMI score of a word pair in a corpus.  

\section{The Universal NLP Engine}

The generic NLP Engine contains (see Figure 1): 

{\bf The transformation engine E}, which takes as input one or more symbols from S and produces zero or more labels in response. 

{\bf The input stream $X$}, which contains the text (without loss of generality, we talk about text (a sequence of words and punctuation), but S might instead be a sequence of symbols from some other vocabulary, such as part of speech tags, or a mixture of several vocabularies, such as words with their individual part of speech tags).  We therefore consider S as consisting of an essentially infinite stream of units, each unit being a symbol (or set of associated symbols) for which a label (or set of labels) is to be computed by E.  Let $\mathbb{X}$ denote the set of source symbols. 

{\bf The data resource(s) R}, typically a lexicon, a grammar, a probability model, or the output of some subtask, used by E to perform its transformation.  

{\bf The output label(s) Y}, a set of predefined symbols that E produces.  Let $\mathbb{Y}$ denote the set of target symbols (including labels). We have $\forall y\in Y, Y\in\mathbb{Y}$ and also possibly $\mathbb{X}\in\mathbb{Y}$.

We next describe a generic procedure for implementing the machine based on the following assumption. 
\begin{Assumption}
 {[}Most] modern NLP tasks can be viewed as predicting a (sequence of) token(s) (i.e., $Y^n$) using a finite-state Turing Machine.
\end{Assumption}
Such a procedure allows one to measure and compare various aspects of [almost] any NLP task and subtask in a systematic way, and to thereby compare the computational properties of alternative approaches and implementations to any NLP (sub)task.  

The following examples, using the same input stream $X^n$=``Dog eats apple", illustrate how the engine works by phrasing several modern NLP tasks as sequential token prediction problems: 
\begin{itemize}
\item Sentiment Classification: \\
 $\mathbb{Y}=\{$``-1", ``0", ``1"$\}$, respectively for negative, neutral, and positive sentiment \\
 $Y^n$=``0" (meaning: neutral sentiment).
\item POS tagging: \\
 $\mathbb{Y}=\{\text{Penn Treebank POS Tags}\}$ \\ 
 $Y_n=$``NNP VBZ NN". 
\item Syntactic Parsing:\\
 $\mathbb{Y}=\{\text{``(ROOT", ``(S", ``(NP", ``)", ...}\}$ \\
 $Y_n$=(ROOT (S (NP (NNP  ) ) (VP (VBZ  ) (NP (NN  ) ) ) ) ). 
\item Semantic analysis:\\
 $\mathbb{Y}=\{\text{English Word List}, \text{Relation List} ...\}$ \\
 $Y^n$=``$\exists$ e . eat ( e ) $\land$ agent ( e , dog ) $\land$ patient ( e , apple )". 
\item Word Sense Disambiguation: \\
 $\mathbb{Y}=\{\text{``1", ``2", ``3", ``4", ...}\}$ \\
 $Y_n$=``1 3 1", correspond to the 1st, 3rd, and 1st senses for the correspondent token.  
\item Machine Translation: \\
 $\mathbb{Y}=\{\text{French Words, Punctuations}\}$ \\
 $Y_n$=``chien mange pomme".  
\item Summarization:\\
 $\mathbb{Y}=\{\text{English Words, Punctuations}\}$ \\
 $Y_n$=``dog eats apple" (the gold-standard summary is the original sentence). 
\end{itemize}

The proposed framework is inspired by recent progress of sequence-to-sequence prediction models in NLP, 
such as machine translation \cite{sutskever2014sequence,bahdanau2014neural,vinyals2014grammar,cho2014learning,graves2014neural} 
and the work of Andreas et al. \shortcite{andreas2013semantic} that illustrates that semantic parsing can to some extent be viewed as a machine translation problem. 
Treating syntactic parsing as a string prediction task is defined and implemented in \cite{vinyals2014grammar}. 

\section{The Universal Entropy Model}

In this section, we discuss the generic model for computing Entropy within a Language Engine.   

\subsection{Requirements}
We first identify requirements for the universal model.  The random guess model satisfies the generality property as it can be 
used in any predictive model.  But it of course does not constitute a good predictive model, since it delivers estimated 
distributions far away from the actual distribution.  
In contrast, n-order Markov models (n=1,2,3...) seem to serve the purpose well and can be easily applied in sequence-labeling 
tasks such as NER and POS tagging.  
However it is tricky to adapt them to single-label predictions such as text classification.  Additionally, their dependence on 
specific feature selections make fair comparison across different implementations complicated or impossible. 

This consideration leads to the first requirement for a model:
\begin{Requirement}
 The model should be able to leverage different types of features {\bf automatically} to avoid infinitely complicated feature engineering procedures. 
\end{Requirement}
We consider $P(Y^n)$ and $P(Y^n|X^n)$ to gain insights about better predictions for entropy calculation.  
We use $P(Y^n)$ for illustration as it can be easily extended to $P(Y^n|X^n)$. 
In the sequence prediction task, let $F_n$ denote the entropy where predictions are made given previous $n$ tokens (an n-order Markov model). 
As proved by Shannon \cite{shannon1951prediction}, $F_n$ monotonically decreases with respect to $n$ and $H(Y^n)$ is strictly bounded by $F_n$:
\begin{equation}
 F_1\geq F_2\geq ...\geq F_{\infty} \geq H(Y^n)
\end{equation}
Taking as example an n-gram word prediction model, theoretically estimated entropy decreases as predictions are made based on increasingly many preceding tokens, roughly stated in \cite{shannon1951prediction}. However, issues arise when $n$ is too large to maintain an n-gram probability table. 

Based Shannon's proof, for the purpose of to the largest extent approximating real entropy using estimated entropy,
we generalize the second property of the model as follows: 
\begin{Requirement}
 The model should be able to memorize earlier information as much as possible given the computing power and storage capacity available. 
\end{Requirement}

\subsection{Model}

This line of thinking suggests using as model recurrent neural networks \cite{mikolov2010recurrent,funahashi1993approximation} or sophisticated versions like LSTM \cite{hochreiter1997long} (Please see Appendix for details about LSTM models).  

Recurrent networks obtain a fixed-sized vector for each step within the processing sequence by convoluting current information with output from earlier step(s).
Such vectors can be viewed as combining evidence obtained so far, and are used to predict the subsequent token(s), typically using a softmax function.  
For labeling tasks, recurrent neural networks first map the input $X^n$ of arbitrary length to a fixed-sized vector, which can be viewed as evidence, and then map that vector to the output by convoluting feature representations at each step. 

Recurrent models have the following merits:
(1) They obey {\bf requirement 1} by automatically encoding ``features" in the real-valued representation vectors without explicit feature selection and engineering. Though the models still require significant parameter tuning, they provide a relatively unified procedure for comparison. 
(2) By sequentially convoluting each token with output from earlier step(s) they have the ability to `remember' information required to approximate (to some degree) the conditional probability of $\lim_{n\rightarrow\infty} P(y_t| y_{t-1}, y_{t-2}, ..., y_{t-n})$, which partially addresses {\bf requirement 2}. 
(3) The model is manageable since it uses constant memory size and runs in linear time.  

We explicitly do not claim that recurrent neural models are a perfect choice as model in an NLP Engine. We acknowledge their numerous shortcomings, and discuss some pros and cons in the concluding section.   
However, we believe that they do offer advantages over other models we have considered with regard to tradeoffs of generality, computing power, and storage capacity. 
These thoughts are inspired by recent progress of sequence-to-sequence generation models \cite{sutskever2014sequence,bahdanau2014neural,vinyals2014grammar,cho2014learning}.

\section{Experiments: Comparing the Uncertainty of NLP Tasks}

Using the above framework, we now calculate the exact entropy for a few NLP tasks.
What must never be overlooked however is the impact of the training/testing datasets used (e.g., the complexity for guessing subsequent words in novels and newspapers can be different) 
and how exactly the task is defined (e.g., differences of complexity in sentiment classification between a 5-class and 2-class problem are huge). 

\subsection{Tasks and Datasets}

\paragraph{Prediction Tasks:}
We use Wikipedia 2014 corpus, divided half and half for training and testing. 
We employ the most-frequent 200,000 words and add an ``unknown" symbol to represent the remainder, making it a 200,001-class prediction problem.  
This is a simple Prediction task. 

\paragraph{Sentiment Analysis}
 \cite{pang2002thumbs}'s dataset comprises sentences containing gold-standard sentiment labels tagged at the start of each sentence. 
We divide the original dataset into training(8101)/dev(500)/testing(2000).  This is an Unaligned Labeling (single) task. 

\paragraph{Question-Answering (UMD)} 
The dataset comprises two domains, History and Literature, and contains roughly 2,000 questions where each question is paired with an answer \cite{iyyer2014neural}.  
Since answers are selected from a pool of roughly 100 answer candidates, this is not an open QA problem but a multi-class classification problem; i.e., 
an Unaligned Labeling (single) task. 

\paragraph{Machine Translation} 
We use the WMT14 English-French dataset and the OpenMT12 English-Chinese dataset.  This is an Unaligned Labeling (sequence) task. 

\paragraph{Part-of-Speech Tagging (Penn Treebank)} 
We use a random sample of Wiki2014 as training and testing dataset, each of which containing 1 million sentences. 
Gold-standard labels are assigned using the Stanford POS tagger. 
This is an Aligned Labeling  task. 
\paragraph{Name Entity Recognition (CoNLL)}
We use the CoNLL-2003 English benchmark for training,
which labels four entity types (person, location, organization, miscellaneous). 
The models are tested on CoNLL-2003 testing data..
This is an Aligned Labeling task.

\paragraph{Syntactic Parsing} 
Training data is the OntoNotes corpus \cite{hovy2006ontonotes} and English Web Treebank \cite{petrov2012overview} with an additional 5 million random sentences, all  parsed by the Stanford Parser \cite{socher2013parsing}.  
The testing dataset is Section 22 of the Penn Treebank plus 1000 sentences from the Question Treebank.  We followed protocols defined in \cite{vinyals2014grammar}.
This is an Unaligned Labeling (sequence) task. 
\paragraph{Question Answer (Open-domain)} 
We use the Yahoo Comprehensive QA dataset. 
The dataset comprises roughly 4 million QA pairs.  Questions and answers are sequences of tokens. 
Questions are treated as inputs and models predict word sequences as responsive answers. 
This is an Unaligned Labeling (sequence) task. 

\subsection{Implementations}
\subsubsection{Prediction Task}
Implementations for prediction tasks, where $P(Y^n)$ is to be estimated, are similar to recurrent language models as defined in \cite{mikolov2010recurrent}. 
Let $e_{t-1}$ denote the representation obtained for timestep $t-1$ based on preceding information from the LSTM. 
Let $e_{Y_t}$ denote the feature representation for the token to be predicted at time $t$. 
By adopting a softmax function, the conditional probability for the occurrence of the current token given earlier evidence is given by:  
\begin{equation}
 p(y_{t}| y_{t-1}, ..., y_{t-n})== \frac{f(e_{t-1},e_{y_t})}{\sum_{Y\in\mathbb{Y}}f(e_{t-1},e_{y})}  
\end{equation}
where $f(e_{t-1},e_{y_t})$ denotes the compositional function between vectors $e_{t-1}$ and $e_{y_t}$.
In this paper, we adopt the form of exponential dot product for $f(\cdot)$:
\begin{equation}
 f(e_{t-1},e_{y_t})=\exp (e_{t-1}\cdot e_{y_t})
\label{eq1}
\end{equation}

\subsubsection{Labeling  Task}
We refer to frameworks \cite{sutskever2014sequence,bahdanau2014neural,vinyals2014grammar}) by first concatenating input and output $\{X^n,Y^n\}=\{x_1, .., x_n, y_1,.., y_n\}$. 
Let $e_{t-1}$ denote the LTSM output at timestep $t-1$ by convoluting all preceding tokens before $t$ in  $\{X^n,Y^n\}$, i.e., $\{x_1,...,x_n,y_1, ..., y_{t-1}\}$. 

\paragraph{Unaligned Single Labeling}
Single-tag Labeling corresponds to the special case where the size of $Y^n$ is 1. 
Taking Sentiment Analysis as an example, sentence-level embeddings (denoted as $e_n$, where $n$ denotes the length of the current sentence) are first obtained recurrently from the LSTM. As it is a binary classification problem, we have:
\begin{equation}
\begin{aligned}
 P(y|\cdot)=\frac{\exp(e_n\cdot e_{y})}{\sum_{y'\in\{0,1\}}\exp(e_n\cdot e_{y'})}
\end{aligned}
\end{equation}
Question-Answering (UMD) is implemented in a similar way. 

\paragraph{Unaligned Sequence Labeling}
Following \cite{bahdanau2014neural,vinyals2014grammar}, the conditional probability for predicting the current token $y_t$ in $Y^n$ is given by 
\begin{equation}
\begin{aligned}
 P(Y^n|X^n)&=\prod_{1\leq t\leq n}P(y_t|x_1,...,x_n,y_1,y_{t-1})\\
 &=\prod_{1\leq t\leq n} \frac{f(e_{t-1},e_{y_t} )}{\sum_{y\in\mathbb{Y}}f(e_{t-1},e_{y})}
\end{aligned}
\end{equation}
$f(\cdot)$ takes the same form as in Eq.\ref{eq1}. 

\paragraph{Aligned Sequence Labeling}
In aligned sequence labeling tasks, there is a one-to-one correspondences between output $y_t$ and input $x_t$, which should be captured in the model. 
Decisions at timestep $t$ are made by combining LSTM representation $e_{t-1}$  and input representation $e_{x_t}$:
\begin{equation}
\begin{aligned}
 P(Y^n|X^n)&=\prod_{1\leq t\leq n}P(y_t|x_1,...,x_n,y_1,...,y_{i-1})\\
 &=\prod_{1\leq t\leq n} \frac{f(e_{t-1},e_{y_t}, e_{x_i} )}{\sum_{y\in\mathbb{Y}}f(e_{t-1},e_{y}, e_{x_i})}
\end{aligned}
\end{equation}
$f(e_{t-1},e_{y_t}, e_{x_i})$ is given as below:
\begin{equation}
\begin{aligned}
 f(e_{t-1},e_{y_t}, e_{x_i})=\exp(U\cdot (W\cdot  [e_{t-1},e_{y_t}, e_{x_i}]))
\end{aligned}
\end{equation}
where $[e_{t-1},e_{y_t}, e_{x_i}]$ denotes the concatenation of the three vectors and $W$ and $U$ denote convolutional matrix and vector to project the concatenated vector to a scalar.
Taking {\bf POS tagging} as example, for the sentence $X_n=$``dog eats bones" with correspondent labels $Y_n=$``NN VBZ NNS", we first concatenate $X_n$ with $Y_n$: ``dog eats bones NN VBZ NNS".
When making predictions at token ``VBZ", let $e_{LSTM}$ denote the LSTM embedding computed at preceding token ``NN", $e_{\text{VBZ}}$ denote the embedding for token ``VBZ", $e_{\text{eats}}$ denote the correspondent input embedding for token ``$\text{eats}$".  Then the probability for generating part-of-speech tag VBZ is given by:
\begin{equation}
 p(\text{VBZ}|\cdot)=\frac{f(e_{LSTM},e_{\text{VBZ}},e_{\text{eats}})}{\sum_{y\in \in\mathbb{Y}}f(e_{LSTM},e_{\text{y}},e_{\text{eats}})}
\end{equation}

\subsection{Details}
For each task, word embeddings are initialized o the same pre-trained vectors for fairness. 
Pre-trained embeddings were obtained from word2vec
on a 6-billion-word corpus with dimensionality 512. 
LSTM models are composed of one single hidden layer. 
Stochastic gradient decent (without momentum) with mini-batch \cite{cotter2011better} is adopted.  
For each task, we use a learning initial learning rate of 0.5 with a linear decay. Learning stops after 4 iterations.  
We initialized the LSTM parameters using a uniform distribution between [-0.1, 0.1].
Referring to \cite{sutskever2014sequence}, the gradient is normalized if its value exceeds a threshold to avoid exploding gradients.
For unaligned sequence prediction tasks (i.e., syntactic parsing, QA(open domain)), inputs are reversed, as suggested in \cite{sutskever2014sequence}.

\subsection{Results}

\begin{table}
\centering
\begin{tabular}{cc}\hline
Task& Avg Entropy\\\hline
Word Prediction (Wiki)& 7.12 \\
English-Chinese Translation&5.17\\
English-French Translation&3.92 \\
QA (Open-domain)&3.87  \\
Syntactic Parsing&1.18\\
QA (UMD)&1.08\\
Text Classification (20 news)&0.70 \\
Sentiment  (Pang)&0.58 \\
Part-of-Speech Tagging& 0.42\\
Name Entity Recognition &0.31\\\hline
\end{tabular}
\caption{Average entropy for different NLP tasks with correspondent dataset specified.}
\label{entropy}
\end{table}

Estimated entropies for different tasks computed in the proposed paradigm are presented in Table~\ref{entropy}. 
As can be seen, MT is less complex than word prediction tasks, which is in line with our expectation: for MT, output tokens are predicated on source tokens. 
The input data provides additional information and lowers the degree of uncertainty: $H(Y|X)\geq H(Y)$ for any $X$ and $Y$. 

As discussed earlier, estimated entropies are subjective to datasets. 
Being significantly short in training data, a high level of entropy is observed for summarization. 
This phenomenon demonstrate one key disadvantage of the proposed model---the failure to consider the impact of datasets. 
In particular, we are computing the upper bound for a specific task given {\bf the specific dataset adopted}.
How to take into account the influence of different datasets (e.g., amounts of training data, quality of training data) poses a great challenge to developing a general NLP Engine. 

\section{Deficiencies and Directions for Improvement}

We have proposed a paradigm with three requirements, which we believe to be essential for a universal NLP engine. 
We are fully aware that it is impossible to come up with instantiations that perfectly meet all the requirements using current algorithms and frameworks. 
We consider the search for optimal solutions to be a long-term task. 
In this section we identify deficiencies involved in the proposed framework and suggest avenues for improvements. 

{\bf The Metric:} We proposed to use Shannon Entropy as uncertainty measurement to evaluate the complexity of tasks because we believe that entropy more deeply reflects the nature of uncertainty than other current measures such as accuracy or recall. However, if a theoretical computer scientist were to develop a more optimal measure that avoids the dilemma described in Section~2, we would replace entropy with that measure. 

{\bf The Engine:} In this paper, we are using an end-to-end turning string prediction engine, which says nothing substantive about complexity of resource and intermediate procedures. This could be problematic. Consider the following scenarios:
in case 1 we have a long table that lists each input possibility and its output answer is a simple lookup, where the work then goes into creating the table, and in case 2 we have a small resource of rules but a lot of feature creation and rule application in the main engine to perform the same task.  It is then true that the entropy from input to output is the same if the two systems produce exactly the same output (though one takes perhaps a lot more time, the other requires perhaps more space).  But is the amount of work required (and hence the entropy effect) to create the two resources the same?  In other words, can one argue that because the ‘outside’ end-to-end
turning prediction
 task is constant in entropy, therefore the inner resources have to contain the same amount of entropy reducing ‘power’?  This is not necessarily true.  But should the one resource contain significantly more than the other, it appears that the outside engine doesn’t actually use that.

{\bf The Model:} Before discussing disadvantages of applied recurrent neural models, it is noteworthy that there is an alternative to a universal and unified model (we call it {\it unified model} for short) for the framework. 
One can instead find the best informants (we call such a strategy {\it best models}) from various places and ask them to perform the transformation predictions. 
Alternatively, one can exhaust all combinations of models, algorithms, and features, and report the best results (smallest value of entropy) as the complexity comparison. 
Though all these strategies have pros and cons, we {\bf postulate} that unified models might be more suitable than best models, as different informants might have 
different levels of education.

To meet the two requirements described in Section~6, we adopted recurrent neural models. 
Recurrent models are by no means perfect: they inevitably forget previous information and are fundamentally incapable of capturing long-term dependencies  \cite{bengio1994learning}. 
This becomes especially problematic in tasks where long-term dependencies play a vital role such as discourse parsing. 
Without trying to defend the model too far, we note that recurrent models seem to offer advantages over other current models that we can think of, e.g., transition models. We are optimistic that other and more sophisticated variations of neural models or other models, such as LDCRFs (Long-Dependency CRFs) \cite{morency2007latent}, memory networks \cite{weston2015towards}
will cope with the aforementioned disadvantages bit by bit.  At least, one can replace recurrent models if more suitable algorithms come up. 

\section{Conclusion: Toward a Theory of NLP}

Almost all NLP researchers today would all agree that there is no such thing as a theory of NLP.  We hope that in this paper we lay some groundwork toward such a theory.  

Any theory addresses some complex phenomenon by (i) identifying some categories (of objects or states or events) within it, (ii) providing some characteristics and perhaps some definitions for them, (iii) if possible describing some relationships between them, and (iv) if possible quantifying (some) aspects of these relationships.  A scientific theory measures aspects of some phenomena and uses rules expressing the relationships to predict the values of other phenomena under certain conditions.  

The framework outlined in this paper names as categories the commonly used linguistic phenomena of NLP such as words, part of speech tags, syntactic classes, and any other linguistically motivated category that NLP researchers choose to study.  But it also has as categories various algorithms and data structures and other aspects of computation, including language models, the notion of training data and evaluation against a gold standard, classification, scoring, etc.  The General NLP Engine puts the notions together in a single generic framework and suggests a way to measure their separate individual characteristics with regard to a single whole, namely the performance of tasks phrased in a very generic manner.  This allows one to hold all but one category constant and vary the characteristics of either a linguistic or a computational category and study its effect on the overall task relative to any other variation, even if applied to some other category.  

It is of course possible to generalize the General NLP Engine to apply to many other application areas in Computer Science.  However the domain of NLP has properties that make it very attractive for fleshing out the nature of the Engine and the general `theory', among others that NLP is a relatively mature domain within Computer Science, being just over 60 years old; NLP addresses a very large and complex subject field, namely natural language, NLP uses a variety of quite different techniques, including finite state transformation engines, machine learning, etc., and numerous types of representations, including vector spaces, symbolic notations, and connectionist embeddings.  

In summary, though far from perfect, this paper provides a first attempt to quantify NLP tasks under a uniform paradigm
which might 
have the potential to
significantly impact natural language processing
areas.

\bibliographystyle{acl}
\bibliography{acl2013}  

\section{Appendix} 
\paragraph{Long-short Term Memory}
LSTM model, first proposed in \cite{hochreiter1997long}, maps an input sequence to a fixed-sized vector by sequentially convoluting the current representation with the output representation of the previous step. LSTM associates each time epoch with an input, control and memory gate, and tries to minimize the impact of unrelated information. Letting $i_t$, $f_t$ and $o_t$ correspond to gate states at time $t$, $e_{t-1}$ and $e_t$ denote the output representation at time $t-1$, and $t$, $e_{x_t}$ denote the embedding associated with the token at time $t$, as defined in \cite{hochreiter1997long}, we have 
\begin{equation}
\begin{aligned}
&i_t=\sigma(W_i\cdot e_{x_t}+V_i\cdot e_{t-1})\\
&f_t=\sigma(W_f\cdot e_{x_t}+V_f\cdot e_{t-1})\\
&o_t=\sigma(W_o\cdot e_{x_t}+V_o\cdot e_{t-1})\\
&l_t=\text{tanh}(W_l\cdot e_{x_t}+V_l\cdot e_{t-1})\\
&m_t=f_t\cdot m_{t-1}+i_t\cdot l_t\\
&e_{t}=o_t\cdot m_t
\end{aligned}
\end{equation}
where $\sigma$ denotes the sigmoid function. $i_t$, $f_t$ and $o_t$ are scalars within the range of [0,1]. 

\end{document}